\documentclass[preprint]{elsarticle}

\usepackage{lineno,hyperref}
\usepackage{amsmath}
\usepackage{mathrsfs}
\usepackage{url}
\usepackage[utf8]{inputenc}
\usepackage{lmodern}
\usepackage{dsfont}
\usepackage{amssymb}
\usepackage{color}
\usepackage{array,booktabs}

\modulolinenumbers[5]

\journal{Neurocomputing}

\begin{document}

\begin{frontmatter}

\title{Cumulative link models for deep ordinal classification}

\author{V\'ictor Manuel Vargas}
\author{Pedro Antonio Guti\'errez}
\author{C\'esar Herv\'as-Mart\'inez}
\address{Department of Computer Science and Numerical Analysis, University of C\'ordoba, C\'ordoba, Spain}
\cortext[a]{V\'ictor Manuel Vargas is the corresponding author.}

\begin{abstract}
	This paper proposes a deep convolutional neural network model for ordinal regression by considering a family of probabilistic ordinal link functions in the output layer. The link functions are those used for cumulative link models, which are traditional statistical linear models based on projecting each pattern into a 1-dimensional space. A set of ordered thresholds splits this space into the different classes of the problem. In our case, the projections are estimated by a non-linear deep neural network. To further improve the results, we combine these ordinal models with a loss function that takes into account the distance between the categories, based on the weighted Kappa index. Three different link functions are studied in the experimental study, and the results are contrasted with statistical analysis. The experiments run over two different ordinal classification problems and the statistical tests confirm that these models improve the results of a nominal model and outperform other robust proposals considered in the literature.
\end{abstract}


\end{frontmatter}


\section{Introduction}
\label{sect:introduction}
Deep learning, introduced by Lecun~et al. \cite{lecun2015deep}, combines multiple machine learning techniques and allows computational models that are composed of numerous processing layers to learn representations of data with various levels of abstraction. These methods have dramatically improved the state-of-the-art in many domains, such as image classification~\cite{yu2017convolutional,fang2019image,zhu2019image}, speech recognition~\cite{song2018music,liu2018speech}, control problems~\cite{mnih2015human}, object detection~\cite{sun2018face,chu2018deep,olmos2018automatic}, privacy and security protection~\cite{tan2018deep,yuan2019adversarial}, recovery of human pose~\cite{hong2015multimodal}, semantic segmentation~\cite{li2019deep} and image retrieval~\cite{tzelepi2018deep,qayyum2017medical,bai2018optimization}. Convolutional Neural Networks (CNN) are one of the types of deep networks that are designed to process data that come in the form of multiple arrays. CNNs are appropriate for images, video, speech and audio processing, and they have been used extensively in the last years for automatic classification tasks~\cite{dong2014learning,ronneberger2015u}. For images, each colour channel is represented by a 2D array, and convolutional layers extract the main features from the pixels, and, after that, a fully connected layer classify every sample based on its extracted features. At the output of the CNN, a softmax function provides the probabilities of the set of classes predefined in the model for classification tasks. However, the softmax may not be the best option depending on the classification problem considered.

Ordinal classification problems are those classification tasks where labels are ordered, and there are different inter-classes importances for each pair of categories. This kind of problem can be treated as nominal classification, but this discards the ordinal information. A better approach is to use specific methods that take the ordinality into account to improve the performance of the classification model. The Proportional Odds Model (POM)~\cite{mccullagh1980regression} is an ordinal alternative to the binary logistic regression. It belongs to a wider family of models called Cumulative Link Models (CLMs)~\cite{agresti2010analysis}. CLMs are inspired in the concept of a latent variable that is projected into a 1-dimensional space and a set of thresholds that divides the projection into the different ordinal levels. A link function needs to be specified, which can be of different types, although the most common option is the \texttt{logit}, which is used in POM. In this paper, we explore different existing alternatives, as explained in depth in Section \ref{sect:ordinalproblem}.

In this paper, we propose the use of CLMs for deriving deep learning ordinal classifiers\footnote{The source code is available at \url{https://github.com/ayrna/deep-ordinal-clm}.}. In the case of CNNs, the model projection used by the threshold model can be obtained from the last layer of the network. Given that we work with a 1-dimensional space, the last layer would have only one neuron (projection of the pattern), and its value could be used to classify the sample into the corresponding class according to the thresholds. Some previous works have used the \texttt{logit} in shallow neural networks~\cite{gutierrez2016ordinal}, but this strategy has not been considered for deep learning, and alternative link functions have not been evaluated. To further improve the results, we train these models by minimising an ordinal loss function based on the Weighted Kappa index \cite{de2018weighted}, instead of using the standard cross-entropy.

An experimental study evaluating the three most common link functions is performed. Also, other parameters that can affect the training process and the model performance are studied, such as the learning rate of the optimization algorithm, the batch size, and their interaction. The nominal version of this model is used as a baseline for comparison. We contrast the results obtained with a statistical analysis to provide more robust conclusions. An ANOVA III test \cite{miller1997beyond}, followed by a posthoc Tukey's test \cite{tukey1949comparing}, is performed over $5$ runs of the experiments, because of the demands of computational time required to run a higher number of executions. The experiments are run using two different ordinal datasets: Diabetic Retinopathy~\cite{de2018weighted}, which contains high-resolution fundus images related with diabetes disease, and Adience~\cite{beckham2017unimodal}, which includes human faces images associated with an age range.

The main contribution of this work is to introduce CLM for CNNs combined with the QWK loss function to achieve high classification performance and better stability than previous works in terms of standard deviation and stagnation problems.

The paper is organized as follows: in Section~\ref{sect:relatedwork}, we analyse previous related works. Section~\ref{sect:modelProposal} presents a formal description of the proposal in this paper, which combines a CLM with an ordinal loss function. In Section~\ref{sect:experiments}, we describe the experiments and the datasets used, while, in Section~\ref{sect:results}, we present the results obtained and the statistical analysis. Finally, Section~\ref{sect:conclusions} exposes the conclusions of this work.

\section{Related works}
\label{sect:relatedwork}
There are many works related to the application and development of CNN models \cite{liu2017deep}, but few works focus on ordinal classification problems. The existing deep ordinal approaches are mainly based on simply using an ordinal evaluation metric, on solving the ordinal problem as multiple binary sub-problems, on using an ordinal loss functions or on constraining the probability distribution of the output layer. These works are described in the following subsections.

\subsection{Simply using an ordinal evaluation metric.}

Alali et al. \cite{alali2018multi} proposed a complex CNN architecture for solving Twitter Sentiment Classification as an ordinal problem. They checked that using average pooling preserves significant features that provide more expressiveness to ordinal scale. They didn't propose any method to include the ordinal information into the classifier, but they tried to find the best CNN model architecture based on an ordinal metric.

\subsection{Solving the ordinal problem as multiple binary sub-problems}

Niu et al. \cite{niu2016ordinal} proposed a learning approach to address ordinal regression problems using CNNs. They divided the problem into a series of binary classification sub-problems and proposed a multiple output CNN optimization algorithm to collectively solve these classification sub-problems, taking into account the correlation between them.

Li et al. \cite{li2017deep} applied deep learning techniques for solving the ordinal problem of Alzheimer's diagnosis and detecting the different levels of the disease as multiple binary sub-problems.

Liu et al. \cite{liu2017deep} proposed a new approach which transforms the ordinal regression problem to binary classification sub-problems and use triplets with instances from different categories to train deep neural networks. In this way, high-level features describing the ordinal relationships are extracted automatically. Given that triplets must be generated, this approach is only recommended for small datasets.

Chen et al. \cite{chen2017using} proposed a deep learning method termed Ranking-CNN. This method combines multiple binary CNNs that are trained with ordinal age labels. The binary outputs are aggregated for the final age prediction. They achieved a tighter error bound for ranking-based age estimation.

In general, all these approaches increment the number of parameters to adjust, as several binary classifiers are simultaneously learnt.

\subsection{Using an ordinal loss function}

De la Torre et al.~\cite{de2018weighted} proposed the use of a continuous version of the quadratic weighted kappa (QWK) metric as loss function for the optimization algorithm. They compared this cost function against the traditional log-loss function using three different datasets, including the Diabetic Retinopathy database as the most complex one. They proved that their function could improve the results as it reduces overfitting and training time. Also, they checked the importance of hyper-parameter tuning. Later, in 2019, another work related to the Diabetic Retinopathy dataset was published \cite{de2019deep} where the authors combined the QWK loss function with the use of images of a higher resolution and new dataset partitions with many more samples on the training split. Although they achieved a proper classification score, the test set only contained a small portion of the test patterns used in the Kaggle competition (around $12\%$ of the samples for testing against $60\%$ on the initial dataset splits).

Rios et al. \cite{rios2017ordinal} presented a CNN model designed to handle ordinal regression tasks on psychiatric notes. They combined an ordinal loss function, a CNN model and conventional feature extraction. Also, the authors applied a technique called Locally Interpretable Model-agnostic Explanation (LIME) to make the non-linear model more interpretable.

Fu et al. \cite{fu2018deep} applied deep learning techniques to Monocular Depth Estimation. They introduced a spacing-increasing discretization strategy to treat the problem as an ordinal regression problem. They improved the performance when training the network with an ordinal regression loss. Also, they used a multi-scale network structure that avoids unnecessary spatial pooling.

Pal et al. \cite{pal2018severity} defined a loss function for CNN that is based on the Earth Mover's Distance and takes into account the ordinal class relationships.

Liu et al. \cite{liu2018constrained} proposed a constrained optimization formulation for the ordinal regression problem which minimizes the negative loglikelihood for a multi-class problem constrained by the order relationship between instances.	

Although the use of these losses introduces the ordinality in model learning, the nature of the models remain nominal.

\subsection{Unimodal probability distributions}

Beckham and Pal~\cite{beckham2017unimodal} proposed a straightforward technique to constrain discrete ordinal probability distributions to be unimodal, via the use of the Poisson and binomial probability distributions. The parameters of these distributions were learnt by using a deep neural network. They evaluated this approach on two large ordinal image datasets, including the Adience dataset used in this paper, obtaining promising results. They also included a simple squared-error reformulation \cite{beckham2016simple} that was sensitive to class ordering.

This approach is the one most related to the CLMs considered in this paper. However, CLMs indirectly model a latent space together with the set of threshold separating the ordered classes, which provides a more flexible and interpretable approach to deep ordinal classification.	

\section{Model proposal}
\label{sect:modelProposal}

Based on the previous analysis of the state-of-the-art, our proposal is to combine a flexible threshold model in the output layer (different forms of a CLM) with an ordinal loss function, in order to better introduce ordinal constraints during learning.

\subsection{Cumulative Link Model (CLM)}
\label{sect:ordinalproblem}
An ordinal classification problem consists in predicting the label $y$ of an input vector $\mathbf{x}$, where $\mathbf{x} \in \mathcal{X} \subseteq \mathds{R}^K$ and $y \in \mathcal{Y}~=~\{\mathcal{C}_1, \mathcal{C}_2, ..., \mathcal{C}_Q\}$, i.e. $\mathbf{x}$ is in a $K$-dimensional input space, and $y$ is in a label space of $Q$ different labels. The objective in an ordinal problem is to find a function $r : \mathcal{X} \rightarrow \mathcal{Y}$ to predict the labels or categories of new patterns, given a training set of $N$ samples, $D = \{(\mathbf{x}_i, y_i), i = 1, ..., N\}$. Labels have a natural ordering in ordinal problems: $\mathcal{C}_1 \prec \mathcal{C}_2 \prec ... \prec \mathcal{C}_Q$. The order between labels gives us the possibility to compare two different elements of $\mathcal{Y}$ by using the relation $\prec$. This is not possible under the nominal classification setting. In regression (where $y \in \mathds{R}$), real values in $\mathds{R}$ can be ordered by the standard $<$ operator, but labels in ordinal regression ($y \in \mathcal{Y}$) do not carry metric information, i.e. the category serves as a qualitative indication of the pattern rather than a quantitative one.

The Proportional Odds Model (POM) arises from a statistical background and is one of the first models designed explicitly for ordinal regression~\cite{mccullagh1980regression}. It dated back to 1980 and is a member of a wider family of models lately recognised as Cumulative Link Models (CLMs)~\cite{agresti2010analysis}. CLMs predict probabilities of groups of contiguous categories, taking the ordinal scale into account. In this way, cumulative probabilities $P(y \prec C_q |\mathbf{x})$ are estimated, which can be directly related to standard probabilities:
\begin{align}
\nonumber P(y \preceq \mathcal{C}_q | \mathbf{x}) &= P(y = \mathcal{C}_1 | \mathbf{x}) + ... + P(y = \mathcal{C}_q | \mathbf{x}),\\
\nonumber P(y = \mathcal{C}_q | \mathbf{x}) &= P(y \preceq \mathcal{C}_q | \mathbf{x}) - P(y \preceq \mathcal{C}_{q-1} | \mathbf{x}),
\end{align}
with $q = 2, ..., Q-1$, and considering that $P(y = \mathcal{C}_1 | \mathbf{x}) = P(y \preceq \mathcal{C}_1 | \mathbf{x})$ and $P(y \preceq \mathcal{C}_Q | \mathbf{x}) = 1$.

The model is inspired by the notion of a latent variable, where $f(\textbf{x})$ represents a one-dimensional mapping. The decision rule $r: \mathcal{X} \rightarrow \mathcal{Y}$ is not fitted directly, but stochastic ordering of space $\mathcal{X}$ is satisfied by the following general model form \cite{herbrich2000large}:
\begin{equation}
\nonumber
g^{-1}(P(y \preceq \mathcal{C}_q | \mathbf{x})) = b_q - f(\mathbf{x}), \quad q = 1, ..., Q-1,
\end{equation}
where $g^{-1} : [0,1] \rightarrow (-\infty, +\infty)$ is a monotonic function often termed as the inverse link function, and $b_q$ is the threshold defined for class $\mathcal{C}_q$. Consider the latent variable $y^* = f(\mathbf{x})^* = f(\mathbf{x}) + \epsilon$, where $\epsilon$ is the random component of the error. The most common choice for the probability distribution of $\epsilon$ is the logistic function (which is the default function for POM). Label $\mathcal{C}_q$ is predicted if and only if $f(\mathbf{x}) \in [b_{q-1}, b_q]$, where the function $f$ and $\mathbf{b} = (b_0, b_1, ..., b_{Q-1}, b_Q)$ are to be determined from the data. It is assumed that $b_0 = -\infty$ and $b_Q = +\infty$, so the real line defined by $f(\textbf{x}), \textbf{x} \in \mathcal{X}$, is divided into $Q$ consecutive intervals. Each interval corresponds to a category. The constraints $b_1 \le b_2 \le ... \le b_{Q-1}$ ensure that $P(y \preceq \mathcal{C}_q | \mathbf{x})$ increases with $q$ \cite{mccullagh1980regression}.

\begin{figure}[!t]
	\centering
	\includegraphics[width=\columnwidth]{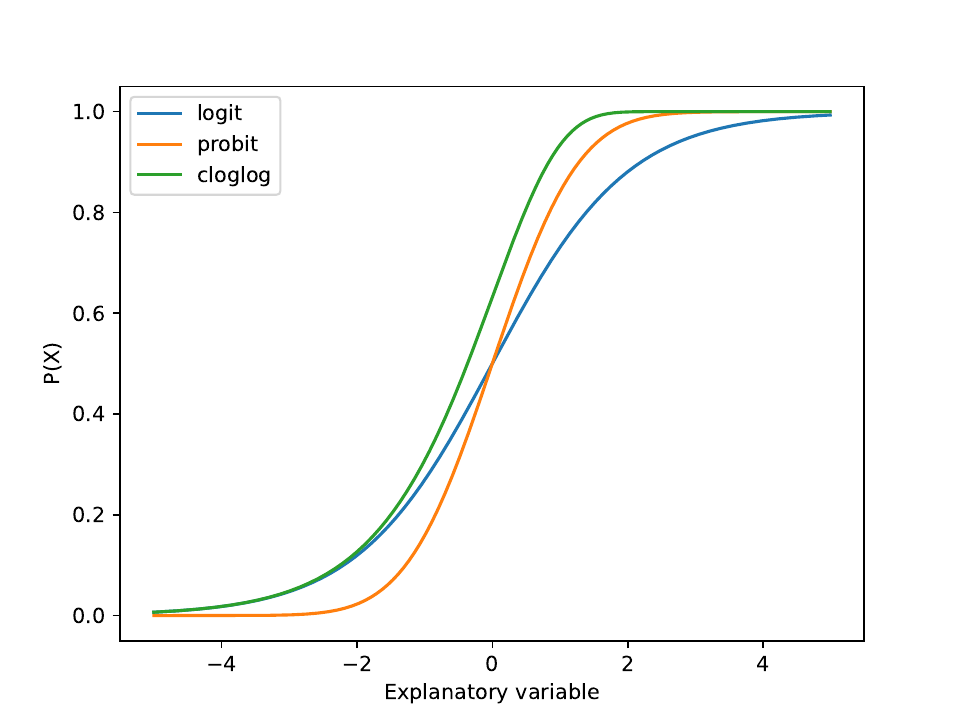}
	\caption{Different link functions commonly used for CLMs.}
	\label{fig:linkfunctions}
\end{figure}

In this work, we consider different link functions previously proposed in CLMs for the probability distribution of $\epsilon$, including \texttt{logit}, \texttt{probit} and complementary log-log (\texttt{clog-log}). These three types of links are explained below and represented in Figure \ref{fig:linkfunctions}. They all follow the same form $\text{link}[P(y \preceq \mathcal{C}_q | \mathbf{x})] = b_q - f(\mathbf{x})$.

\begin{itemize}
	\item The \texttt{logit} link function is the function used for the POM and is defined as:
	\begin{equation}
	\nonumber
	\begin{aligned}
	\text{logit}[P(y \preceq \mathcal{C}_q | \mathbf{x})] = \log\frac{P(y \preceq \mathcal{C}_q | \mathbf{x})}{1 - P(y \preceq \mathcal{C}_q | \mathbf{x})}=& \\ = b_q - f(\mathbf{x}), \quad q = 1, ..., Q-1,
	\end{aligned}
	\label{eq:logit}
	\end{equation}		
	or the equivalent expression:		
	\begin{equation}
	\nonumber
	P(y \preceq \mathcal{C}_q | \mathbf{x}) = \frac{1}{1 + e^{-(b_q - f(\mathbf{x}))}}.
	\label{eq:logit2}
	\end{equation}
	
	\item The \texttt{probit} link function is the inverse of the standard normal cumulative distribution function (cdf) $\Phi$. Its expression is:
	\begin{equation}
	\nonumber
	\begin{aligned}
	\Phi^{-1}[P(y \preceq \mathcal{C}_q | \mathbf{x})] = b_q - f(\mathbf{x}), \quad &q = 1, ..., Q-1,\\
	P(y \preceq \mathcal{C}_q | \mathbf{x}) = \Phi(b_q - f(\mathbf{x})), \quad &q = 1, ..., Q-1,
	\end{aligned}
	\label{eq:probit}
	\end{equation}		
	which can also be expressed as:
	\begin{equation}
	\nonumber
	P(y \preceq \mathcal{C}_q | \mathbf{x}) = \int_{-\infty}^{b_q - f(\mathbf{x})} \frac{1}{\sqrt{2\pi}} e^{-\frac{1}{2}x^2} \mathrm{d}x.
	\label{eq:probit2}
	\end{equation}
	
	\item The \texttt{clog-log} takes a response that is restricted to the $(0,1)$ interval and converts it into a value in the $(-\infty, +\infty)$ interval (like \texttt{logit} and \texttt{probit} transformations). The \texttt{clog-log} expression is:
	\begin{equation}
	\nonumber
	\begin{aligned}
	\log[-\log[1 - P(y \preceq \mathcal{C}_q | \mathbf{x})]] =b_q - f(\mathbf{x}),
	\end{aligned}
	\label{eq:cloglog}
	\end{equation}
	with $q = 1, ..., Q-1,$ that is:
	\begin{equation}
	\nonumber
	P(y \preceq \mathcal{C}_q | \mathbf{x}) = 1 - e^{-e^{b_q - f(\mathbf{x})}}, \quad q = 1, ..., Q-1.
	\label{eq:cloglog2}
	\end{equation}
\end{itemize}

\texttt{logit} and \texttt{probit} links are symmetric:	
\begin{equation}
\nonumber
\text{link}[P(y \preceq \mathcal{C}_q | \mathbf{x})] = -\text{link}[1 - P(y \preceq \mathcal{C}_q | \mathbf{x})],
\end{equation}
which means that the response curve for $P(y \preceq \mathcal{C}_q | \mathbf{x})$ is symmetric around the point $P(y \preceq \mathcal{C}_q | \mathbf{x}) = 0.5$, i.e. $P(y \preceq \mathcal{C}_q | \mathbf{x})$ has the same rate when approaching 0 than when approaching 1. This symmetry property can be demonstrated as follows:	
\begin{enumerate}
	\item Let $P(y \preceq \mathcal{C}_q | \mathbf{x}) \equiv p$. For the \texttt{logit} function, we have:
	\begin{equation}
	\nonumber
	\begin{aligned}
	\text{link}(p) = \text{logit}(p) &= \log\left(\frac{p}{1-p}\right) =\\
	&= \log(p) - \log(1-p),
	\end{aligned}
	\end{equation}			
	while:			
	\begin{equation}
	\nonumber
	\begin{aligned}
	-\text{link}(1 - p) &= -\text{logit}(1 - p) =\\
	=- \log\left(\frac{1- p}{p}\right) &= - \log(1 - p) + \log(p).
	\end{aligned}
	\end{equation}
	
	\item For the \texttt{probit}:		
	\begin{equation}
	\nonumber
	\begin{aligned}
	p \equiv P(y \preceq \mathcal{C}_q | \mathbf{x}) &= \Phi(b_q - f(\mathbf{x})) =\\
	&= \int_{-\infty}^{b_q - f(\mathbf{x})} \frac{1}{\sqrt{2\pi}} e^{-\frac{1}{2}x^2} \mathrm{d}x,
	\end{aligned}
	\end{equation}
	which leads to:
	\begin{align}
	\nonumber \Phi^{-1}(p) & = \Phi^{-1}(p) = b_q - f(\mathbf{x}),\\
	\nonumber -\Phi^{-1}(1-p) &= \Phi^{-1}(1-p) = -b_q + f(\mathbf{x}),
	\end{align}
	where:		
	\begin{equation}
	\nonumber
	1 - p = \int_{-\infty}^{-b_q + f(\mathbf{x})} \frac{1}{\sqrt{2\pi}} e^{-\frac{1}{2}x^2} \mathrm{d}x,
	\end{equation}
	\begin{equation}
	\nonumber
	p = 1 - \int_{-\infty}^{-b_q + f(\mathbf{x})} \frac{1}{\sqrt{2\pi}} e^{-\frac{1}{2}x^2} \mathrm{d}x.
	\end{equation}
\end{enumerate}

Unlike \texttt{logit} and \texttt{probit}, the \texttt{clog-log} link is asymmetrical. In this way, when the distribution of the given data is not symmetric in the $[0,1]$ interval and increase slowly at small to moderate value but increases sharply near 1, the \texttt{logit} and \texttt{probit} models are inappropriate, while \texttt{clog-log} can lead to better results.

In this paper, the probabilistic structure of CLMs is proposed as a link function for deep convolutional neural networks. This can be achieved by defining a new type of output layer alternative to the standard softmax layer. In this way, the proposed output layer will transform the one-dimensional projection, previously denoted as $f(\mathbf{x})$, into a set of probabilities. $f(\mathbf{x})$ is estimated from a nonlinear transformation of the set of features learnt by the previous layers, $l(\mathbf{x})=f(\mathbf{x})$, where $\mathbf{x}$ is the pattern being evaluated and $l(\mathbf{x})$ is a latent representation of the pattern given by the output of a single neuron. In order to apply unconstrained optimizers while ensuring $b_1 \le b_2 \le ... \le b_{Q-1}$, we can redefine the thresholds. All of them can be derived from the first one in the following form:
\begin{equation}
\nonumber
b_q = b_1 + \sum_{q=1}^{q-1} \alpha_q^2, \quad q = 2, ..., Q,
\end{equation}
where $b_1$ is a learning parameter corresponding to the first threshold, $\alpha_q$ is a learnable vector of parameters used to obtain the rest of the thresholds, and $Q$ is the number of classes.

\subsection{Continuous Quadratic Weighted Kappa (QWK) loss function}
\label{sect:wk}

In order to increase the performance of the deep ordinal model, the CLM structure in the output layer is combined with the continuous version of the QWK loss \cite{de2018weighted} function. The Kappa index is a well-known metric that measures the agreement between two different raters. The Weighted Kappa (WK)~\cite{ben2008comparison} is based on the Kappa index and adds different weights to the different types of disagreements based on a weight matrix. It is useful to evaluate the performance in ordinal problems, as it gives a higher weight to the errors that are further from the correct class. This metric is defined as follows:
\begin{equation}
\label{eq:qwk}
\text{QWK} = 1 - \frac{\sum\limits^N_{i,j} \omega_{i,j} O_{i,j}}{\sum\limits^N_{i,j} \omega_{i,j} E_{i,j}},
\end{equation}
where $N$ is the number of samples rated, $\omega$ is the penalization matrix (in this case, quadratic weights are considered, $\omega_{i,j} = \frac{(i-j)^2}{(C-1)^2}$, $\omega_{i,j} \in [0,1]$), $O$ is the confusion matrix, $E_{ij} = \frac{O_{i\bullet} O_{\bullet j}}{N}$, $O_{i\bullet}$ is the sum of the $i\text{-th}$ row and $O_{\bullet j}$ is the sum of the $j\text{-th}$ column.

The WK defined above cannot be used as a loss function for the optimization algorithm as it is not continuous. However, it has been previously redefined \cite{de2018weighted} in terms of probabilities of the predictions:
\begin{equation}
\nonumber
\text{QWK}_c = \frac{\sum\limits_{k=1}^N \sum\limits_{q=1}^Q \omega_{t_k, q} P(y = \mathcal{C}_q | \mathbf{x}_k)}{\sum\limits_{i=1}^Q \frac{N_i}{N} \sum\limits_{j=1}^Q ( \omega_{i,j} \sum\limits_{k=1}^N P(y = \mathcal{C}_j | \mathbf{x}_k))},
\end{equation}
where $\text{QWK}_c \in [0,2]$, $\mathbf{x}_k$ and $t_k$ are the input data and the real class of the $k$-th sample, $Q$ is the number of classes, $N$ is the number of samples, $N_i$ is the number of samples of the $i$-th class, $P(y = \mathcal{C}_q | \mathbf{x}_k)$ is the probability that the $k$-th sample belongs to class $\mathcal{C}_q$ (estimated using the CLM structure), and $\omega_{i,j}$ are the elements of the penalization matrix ($\omega_{i,j} = \frac{(i-j)^2}{(C-1)^2}$). This loss function can be minimized using a gradient descent based algorithm.

\section{Experiments}
\label{sect:experiments}
\subsection{Data}
In order to evaluate the different models, we make use of two ordinal datasets:
\subsubsection{Diabetic Retinopathy (DR)}
\begin{figure}[!t]
	\centering
	\resizebox{\columnwidth}{!}{
	\begin{tabular}{cccc}
		\includegraphics[width=2.4cm]{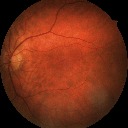} & \includegraphics[width=2.4cm]{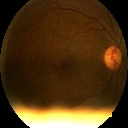} & \includegraphics[width=2.4cm]{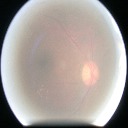} & \includegraphics[width=2.4cm]{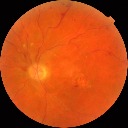}\\
		\includegraphics[width=2.4cm]{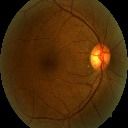} & \includegraphics[width=2.4cm]{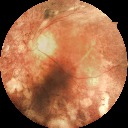} & \includegraphics[width=2.4cm]{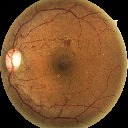} & \includegraphics[width=2.4cm]{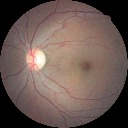}
	\end{tabular}}					
	\caption{Examples taken from the Diabetic Retinopathy test set.}
	\label{fig:DRexamples}
\end{figure}

DR\footnote{https://www.kaggle.com/c/diabetic-retinopathy-detection/data} is a dataset consisting of extremely high-resolution fundus image data. The training set consists of $17563$ pairs of images (where a  pair includes a left and right eye image corresponding to a patient). In this dataset, we try to predict the correct category from five levels of diabetic retinopathy: no DR ($25810$ images), mild DR ($2443$ images), moderate DR ($5292$ images), severe DR ($873$ images), or proliferative DR ($708$ images). The test set contains $26788$ pairs of images. These images are taken in variable conditions: by different cameras,  conditions of illumination and resolutions. They come from the EyePACS dataset that was used in the DR detection competition hosted on the Kaggle platform. Also, this dataset has been used in different works~\cite{de2018weighted,nebot2016diabetic}, where the $\text{QWK}_c$ cost function was considered in \cite{de2018weighted} to achieve better performance. A validation set is set aside, consisting of $10\%$ of the patients in the training set. The images are resized to 128 by 128 pixels and rescaled to $[0,1]$ range. Data augmentation techniques, described in Section \ref{sect:settings}, are applied to achieve a higher number of samples. A few test images of this dataset are shown in Figure \ref{fig:DRexamples}.

\subsubsection{Adience}
\begin{figure}[!t]
	\centering
	\resizebox{\columnwidth}{!}{
	\begin{tabular}{cccc}
		\includegraphics[width=2.4cm]{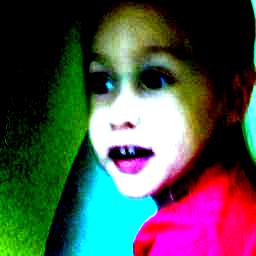} & \includegraphics[width=2.4cm]{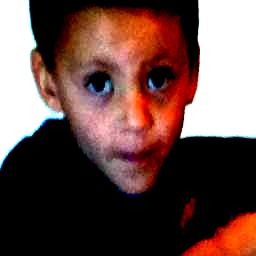} & \includegraphics[width=2.4cm]{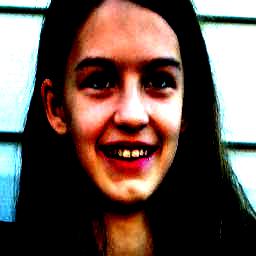} & \includegraphics[width=2.4cm]{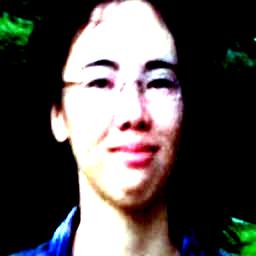}\\
		\includegraphics[width=2.4cm]{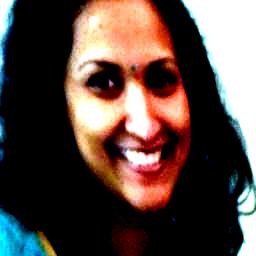} & \includegraphics[width=2.4cm]{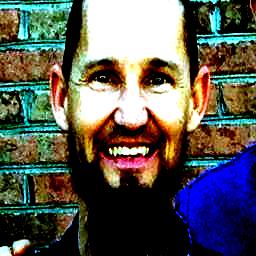} & \includegraphics[width=2.4cm]{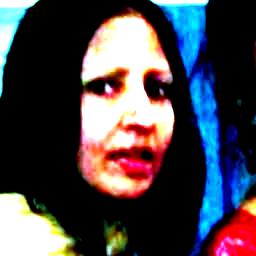} & \includegraphics[width=2.4cm]{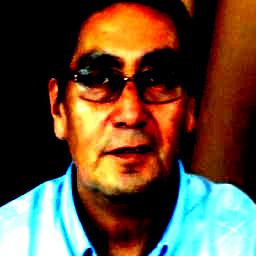}
	\end{tabular}}				
	\caption{Examples taken from the Adience test set.}
	\label{fig:AdienceExamples}
\end{figure}

Adience\footnote{http://www.openu.ac.il/home/hassner/Adience/data.html} dataset consists of $26580$ faces belonging to $2284$ subjects. We use the form of the dataset where faces have been pre-cropped and aligned. The dataset was preprocessed, using the methods described in a previous work~\cite{beckham2017unimodal}, so that the images are 256 pixels in width and height, and pixels values follow a $(0;1)$ normal distribution. The original dataset was split into five cross-validation folds. The training set consists of merging the first four folds which comprise a total of $15554$ images. From this, $10\%$ of the images are held out as part of a validation set. The last fold is used as test set. Some images of this dataset are shown in Figure \ref{fig:AdienceExamples}. Adience dataset has been used in other works for human age estimation but most of them solved the problem as a multi-class problem instead of using the ordinal relation between classes. Eidinger et al. \cite{eidinger2014age} presented an approach using support vector machines and neural networks. Chen et al. \cite{chen2016cascaded} proposed a coarse-to-fine strategy for deep CNNs. Levi and Hassner \cite{levi2015age} presented another convolutional network model for age estimation. As previously discussed, Beckham and Pal \cite{beckham2017unimodal} proposed a straightforward method to constrain discrete probability distributions to be unimodal. M. Duan et al. \cite{duan2018hybrid} proposed a hybrid approach that combines CNN with Extreme Learning Machine (ELM).

\subsection{Model}
\label{sect:model}
CNNs have been used for both datasets. The different architectures of CNNs used in these experiments are presented in Table \ref{table:Architecture}. The architecture for DR is the same that was used in \cite{de2018weighted} and the network for Adience is a small Residual Network (ResNet) \cite{he2016deep} that was used in \cite{beckham2017unimodal}. The most important parameters for convolutional layers are the number of filters that are used to make the convolution operation, the size of these filters and the stride, which is the number of pixels that the filter is moved in every operation. Pooling layers have similar parameters: pool size (number of pixels that will be involved in the operation) and stride. For convolutional layers, ConvWxH@FsS stands for filters of size WxH and stride S. For pooling layers, PoolWxHsS corresponds to a pool size of WxH and stride S.

The Exponential Linear Unit (ELU)~\cite{clevert2015fast} has been used as the activation function for all the convolutional and dense layers, instead of the ReLU~\cite{nair2010rectified} function, as it mitigates the effects of the vanishing gradient problem \cite{bengio1994learning,pascanu2013difficulty} via the identity for positive values. Also, ELUs lead to faster training and better generalization performance than ReLU and Leaky ReLU (LReLU)~\cite{maas2013rectifier} functions on networks with more than five layers.

After every ELU activation function of the convolutional layers, Batch Normalization~\cite{ioffe2015batch} is applied. This method reduces the internal covariate shift by normalizing layer outputs. As the authors stated \cite{ioffe2015batch}, it allows us to use higher learning rates and be less careful about weight initialization. In some cases, it also eliminates the need for using regularization techniques like Dropout.

At the output of the network, the CLM is used (see Section \ref{sect:ordinalproblem}). Also, a learnable parameter $\tau$ rescales the projections used by the CLM to make it more stable and guarantee the convergence in most cases. The following expression describes the transformation applied to these projections:
\begin{equation}
\nonumber
f(\mathbf{x}) = \frac{l(\mathbf{x})}{\tau}.
\end{equation}
where $\tau$ is optimized as a free parameter along with the parameters of the model.

\begin{table}[!t]
	\caption{Description of the architecture used for each dataset.}
	\label{table:Architecture}
	\centering
	\footnotesize
	\resizebox{\columnwidth}{!}{
	\begin{tabular}{lclc}
		\hline\hline
		\multicolumn{2}{c}{\textbf{Diabetic Retinopathy}} & \multicolumn{2}{c}{\textbf{Adience}}  \\ \hline
		\textbf{Layer}  &         \textbf{Output}         & \textbf{Layer}      & \textbf{Output} \\ \hline
		2 Conv3x3@32s1  &           124x124x32            & Conv7x7@32s2        &   112x112x32    \\
		MaxPool2x2s2    &            62x62x32             & MaxPool3x3s2        &    55x55x32     \\
		2 Conv3x3@64s1  &            58x58x64             & 2 ResBlock3x3@64s1  &    55x55x32     \\
		MaxPool2x2s2    &            29x29x64             & ResBlock3x3@128s2   &    28x28x64     \\
		2 Conv3x3@128s1 &            25x25x128            & 2 ResBlock3x3@128s1 &    28x28x64     \\
		MaxPool2x2s2    &            12x12x128            & 1 ResBlock3x3@256s2 &    14x14x128    \\
		2 Conv3x3@128s1 &             8x8x128             & 2 ResBlock3x3@256s1 &    14x14x128    \\
		MaxPool2x2s2    &             4x4x128             & 1 ResBlock3x3@512s2 &     7x7x256     \\
		Conv4x4@128s1   &             1x1x128             & 2 ResBlock3x3@512s1 &     7x7x256     \\
		                &                                 & AveragePool7x7s2    &     1x1x256     \\ \hline\hline
	\end{tabular}}
\end{table}

\subsection{Experimental design}
\label{sect:settings}
Weights are adjusted using a batch based first-order optimization algorithm called Adam~\cite{kingma2014adam}. We study different initial learning rates ($\eta$) in order to find the optimal one for each problem. We apply an exponential decay \cite{chin2015learning} across training epochs to the initial learning rate ($\eta_0$) following the expression below:
\begin{equation}
\nonumber
\eta = \eta_0 \cdot e^{-0.025 \cdot \text{epoch}}.
\end{equation}

Both datasets are artificially balanced using data augmentation techniques~\cite{van2001art}. However, different transformations are applied to each one. DR dataset augmentation is based on image cropping and zooming, horizontal and vertical flipping, brightness adjustment and random rotations. Horizontal flipping is the only transformation applied to the Adience dataset. These transformations are applied every time a new batch is loaded, and the parameters of each one are randomly chosen from a defined range ($[0.8, 1.2]$ for zooming, $[0.5, 1.5]$ for brightness and $[0, 90]$ degrees for rotation), providing a new set of transformed images for each batch. This technique reduces the overfitting risk and provides an important performance boost as we always work with different but similar images \cite{krizhevsky2012imagenet}.

The epoch size is equal to the number of images in the training set. It could be a higher number as we are using data augmentation, but instead of increasing the epoch size, we rather run the training for more epochs. In this case, we set the maximum number of epochs to $100$. However, we always save the best model, that is evaluated when the training finishes.

The models are mainly evaluated using the QWK metric defined in Eq. \eqref{eq:qwk}. Also, other evaluation metrics are used to ease the comparison with alternative works:
\begin{itemize}
	\item Minimum Sensitivity (MS) \cite{cruz2014metrics} is the lowest percentage of samples correctly predicted to belong to a class with respect to the number of samples of that class.
	\begin{equation}
	\nonumber
	\text{MS} = \min\left\lbrace S_q = \frac{O_{qq}}{O_{q\bullet}}, q = 1, ..., Q \right\rbrace,
	\end{equation}
	where $O$ is the confusion matrix and $Q$ is the number of classes.
	
	\item Mean Absolute Error (MAE) \cite{cruz2014metrics} is the average absolute deviation of the predicted category from the real one.
	\begin{equation}
	\nonumber
	\text{MAE} = \frac{1}{N} \sum^Q_{i,j = 1} |i-j|O_{ij},
	\end{equation}
	where $N$ is the number of samples, $Q$ is the number of classes and O is the confusion matrix.
	
	\item Accuracy-based metrics. Correct Classification Rate (CCR) or standard accuracy is the most common metric for classification tasks and shows the percentage of correctly classified samples. We also include Top-2 CCR and Top-3 CCR \cite{beckham2017unimodal}, which are similar to CCR, but they take a prediction as correct when the real class is between the two or three classes, respectively, with the highest probability.
	
	\item 1-off accuracy \cite{eidinger2014age,chen2016cascaded,levi2015age} marks the prediction as correct when the correct class is at one category of distance (in the ordinal scale) from the predicted one.
\end{itemize}
QWK, MAE and 1-off accuracy are ordinal evaluation metrics, while MS, CCR, Top-2 CCR and Top-3 CCR do not take the order of categories into consideration.

In order to ease reproducibility, the source code is available in a public repository\footnote{\url{https://github.com/ayrna/deep-ordinal-clm}}.
\subsection{Factors}
In our study, three different factors are considered:
\begin{itemize}
	\item \textit{Learning rate} (LR, $\eta$). LR is one of the most critical hyper-parameters to tune for training deep neural networks. Optimal learning rate can vary depending on the dataset and the CNN architecture. Previous works have presented some techniques that adjust this parameter in order to achieve better performance~\cite{smith2017cyclical,senior2013empirical}. In this work, we consider three different values for the initial value of this parameter: $10^{-4}$, $10^{-3}$ and $10^{-2}$.
	\item \textit{Batch size} (BS). Batch size is also an important parameter as it controls the number of weight updates that are made on every epoch. It can affect the training time and the model performance. In this paper, we try three different batch sizes for each dataset. For the DR dataset, we use $5$, $10$ and $15$, while, for Adience, $64$, $128$ and $256$ images are used. We took the batch sizes that were used in \cite{de2018weighted} and \cite{beckham2017unimodal} as a reference, and we expand the range on both sides.
	\item \textit{Link function} (LF). Different link functions are used for the CLM at the last layer output: \texttt{logit}, \texttt{probit} and complementary log-log (see Section \ref{sect:ordinalproblem}).
\end{itemize}

\section{Results}
\label{sect:results}
In this Section, we present the results of the experiments. First, in Sections \ref{sect:dr}, \ref{sect:adience}, and \ref{sect:statisticalanalysis}, we perform the study for adjusting the value of the different parameters. Then, Section \ref{sect:NominalComparison} compares the results against the state of the art.

For each dataset, we show a table with the detailed results of the experiments performed for training the model with each combination of parameters. Every parameter combination was run five times. These tables show the mean value and the standard deviation (SD) of each metric across these five executions for the test set.

\subsection{Diabetic Retinopathy}
\label{sect:dr}
Detailed test results for the DR dataset are presented in Table \ref{table:DRresults}. The best result for each metric is marked in bold and the second best is in italic font.

\begin{table*}[!t]
	\caption{DR results. BS stands for Batch Size, LF for link function and LR for Learning Rate. Mean and standard deviation are represented as $\text{Mean}_\text{SD}$.}
	\label{table:DRresults}
	\footnotesize
	\centering
	\resizebox{\linewidth}{!}{
	\begin{tabular}{c@{\hskip 0.15cm}c@{\hskip 0.15cm}c@{\hskip 0.15cm}c@{\hskip 0.15cm}c@{\hskip 0.15cm}c@{\hskip 0.15cm}c@{\hskip 0.15cm}c@{\hskip 0.15cm}c@{\hskip 0.15cm}c}
		\hline
		\hline
		BS & LF & LR & $\overline{\text{QWK}}_{{SD}}$ & $\overline{\text{MS}}_{{SD}}$ & $\overline{\text{MAE}}_{{SD}}$ & $\overline{\text{CCR}}_{{SD}}$ & $\overline{\text{Top-2}}_{{SD}}$ & $\overline{\text{Top-3}}_{{SD}}$ & $\overline{\text{1-off}}_{{SD}}$\\\hline
		5 & \texttt{clog-log} & $10^{-2}$ & $0.414_{0.057}$ & $0.075_{0.042}$ & $0.177_{0.023}$ & $0.556_{0.057}$ & $0.833_{0.042}$ & $0.968_{0.011}$ & $0.816_{0.021}$\\
		5 & \texttt{clog-log} & $10^{-3}$ & $0.534_{0.027}$ & $0.102_{0.011}$ & $0.137_{0.006}$ & $0.658_{0.015}$ & $0.871_{0.011}$ & $0.966_{0.003}$ & $0.852_{0.002}$\\
		5 & \texttt{clog-log} & $10^{-4}$ & $0.520_{0.006}$ & $0.067_{0.008}$ & $0.123_{0.003}$ & $0.697_{0.006}$ & $0.842_{0.008}$ & $0.961_{0.003}$ & $0.851_{0.002}$\\
		5 & \texttt{logit} & $10^{-2}$ & $0.416_{0.041}$ & $0.095_{0.029}$ & $0.175_{0.021}$ & $0.563_{0.054}$ & $0.762_{0.040}$ & $0.908_{0.026}$ & $0.807_{0.029}$\\
		5 & \texttt{logit} & $10^{-3}$ & $0.554_{0.013}$ & $0.093_{0.009}$ & $0.137_{0.003}$ & $0.660_{0.008}$ & $0.802_{0.005}$ & $0.936_{0.004}$ & $0.853_{0.005}$\\
		5 & \texttt{logit} & $10^{-4}$ & $0.520_{0.003}$ & $0.063_{0.004}$ & $0.122_{0.002}$ & $0.706_{0.005}$ & $0.823_{0.004}$ & $0.949_{0.003}$ & $0.862_{0.003}$\\
		5 & \texttt{probit} & $10^{-2}$ & $0.460_{0.048}$ & $0.079_{0.046}$ & $0.197_{0.064}$ & $0.504_{0.167}$ & $0.808_{0.034}$ & $0.927_{0.073}$ & $0.689_{0.240}$\\
		5 & \texttt{probit} & $10^{-3}$ & $0.564_{0.018}$ & $0.099_{0.013}$ & $0.147_{0.018}$ & $0.636_{0.045}$ & $0.822_{0.040}$ & $0.939_{0.020}$ & $0.840_{0.015}$\\
		5 & \texttt{probit} & $10^{-4}$ & $0.523_{0.005}$ & $0.067_{0.012}$ & $0.122_{0.002}$ & $0.701_{0.006}$ & $0.823_{0.002}$ & $0.953_{0.002}$ & $0.860_{0.003}$\\
		10 & \texttt{clog-log} & $10^{-2}$ & $0.423_{0.239}$ & $0.062_{0.051}$ & $0.127_{0.017}$ & $0.684_{0.046}$ & $\mathbf{0.894_{0.062}}$ & $\mathbf{0.986_{0.012}}$ & $0.832_{0.020}$\\
		10 & \texttt{clog-log} & $10^{-3}$ & $\mathbf{0.582_{0.016}}$ & $0.102_{0.006}$ & $0.128_{0.003}$ & $0.680_{0.007}$ & $\mathit{0.880_{0.004}}$ & $0.972_{0.003}$ & $0.861_{0.004}$\\
		10 & \texttt{clog-log} & $10^{-4}$ & $0.537_{0.010}$ & $0.064_{0.004}$ & $\mathit{0.116_{0.001}}$ & $0.717_{0.003}$ & $0.837_{0.002}$ & $0.971_{0.001}$ & $0.860_{0.002}$\\
		10 & \texttt{logit} & $10^{-2}$ & $0.531_{0.031}$ & $0.107_{0.008}$ & $0.151_{0.010}$ & $0.623_{0.025}$ & $0.802_{0.022}$ & $0.934_{0.013}$ & $0.838_{0.014}$\\
		10 & \texttt{logit} & $10^{-3}$ & $0.579_{0.009}$ & $0.096_{0.012}$ & $0.127_{0.005}$ & $0.686_{0.013}$ & $0.817_{0.006}$ & $0.954_{0.005}$ & $0.861_{0.002}$\\
		10 & \texttt{logit} & $10^{-4}$ & $0.539_{0.007}$ & $0.074_{0.013}$ & $0.126_{0.005}$ & $0.707_{0.010}$ & $0.823_{0.007}$ & $0.957_{0.005}$ & $0.858_{0.004}$\\
		10 & \texttt{probit} & $10^{-2}$ & $0.508_{0.037}$ & $0.088_{0.044}$ & $0.145_{0.018}$ & $0.639_{0.045}$ & $0.835_{0.015}$ & $0.960_{0.008}$ & $0.829_{0.020}$\\
		10 & \texttt{probit} & $10^{-3}$ & $0.558_{0.034}$ & $\mathbf{0.111_{0.005}}$ & $0.134_{0.003}$ & $0.666_{0.008}$ & $0.831_{0.007}$ & $0.955_{0.001}$ & $0.863_{0.003}$\\
		10 & \texttt{probit} & $10^{-4}$ & $0.541_{0.010}$ & $0.076_{0.006}$ & $0.119_{0.002}$ & $0.712_{0.005}$ & $0.828_{0.003}$ & $0.961_{0.002}$ & $0.862_{0.001}$\\
		15 & \texttt{clog-log} & $10^{-2}$ & $0.564_{0.016}$ & $0.108_{0.014}$ & $0.143_{0.006}$ & $0.640_{0.015}$ & $0.879_{0.011}$ & $0.972_{0.005}$ & $0.851_{0.006}$\\
		15 & \texttt{clog-log} & $10^{-3}$ & $0.559_{0.026}$ & $\mathit{0.111_{0.008}}$ & $0.127_{0.004}$ & $0.682_{0.010}$ & $0.871_{0.008}$ & $\mathit{0.974_{0.002}}$ & $\mathbf{0.868_{0.002}}$\\
		15 & \texttt{clog-log} & $10^{-4}$ & $0.538_{0.009}$ & $0.054_{0.003}$ & $\mathbf{0.115_{0.002}}$ & $0.720_{0.006}$ & $0.835_{0.007}$ & $0.970_{0.003}$ & $0.860_{0.006}$\\
		15 & \texttt{logit} & $10^{-2}$ & $0.551_{0.020}$ & $0.104_{0.008}$ & $0.139_{0.011}$ & $0.654_{0.027}$ & $0.815_{0.017}$ & $0.948_{0.016}$ & $0.856_{0.015}$\\
		15 & \texttt{logit} & $10^{-3}$ & $0.551_{0.010}$ & $0.106_{0.016}$ & $0.129_{0.008}$ & $0.680_{0.019}$ & $0.818_{0.008}$ & $0.952_{0.007}$ & $\mathit{0.866_{0.001}}$\\
		15 & \texttt{logit} & $10^{-4}$ & $0.543_{0.008}$ & $0.056_{0.003}$ & $0.121_{0.004}$ & $\mathbf{0.723_{0.004}}$ & $0.833_{0.004}$ & $0.964_{0.003}$ & $0.862_{0.004}$\\
		15 & \texttt{probit} & $10^{-2}$ & $0.534_{0.032}$ & $0.104_{0.013}$ & $0.148_{0.015}$ & $0.631_{0.038}$ & $0.845_{0.030}$ & $0.964_{0.010}$ & $0.852_{0.010}$\\
		15 & \texttt{probit} & $10^{-3}$ & $\mathit{0.580_{0.021}}$ & $0.104_{0.016}$ & $0.129_{0.008}$ & $0.680_{0.018}$ & $0.832_{0.010}$ & $0.959_{0.007}$ & $0.866_{0.003}$\\
		15 & \texttt{probit} & $10^{-4}$ & $0.533_{0.004}$ & $0.065_{0.005}$ & $0.117_{0.002}$ & $\mathit{0.721_{0.004}}$ & $0.832_{0.002}$ & $0.964_{0.001}$ & $0.863_{0.001}$\\
		\hline
		\hline
	\end{tabular}}
\end{table*}

The best mean QWK value was obtained with the \texttt{clog-log} link function using a BS of 10 and a LR of $10^{-3}$. However, the best CCR value was obtained with a BS of 15, the \texttt{logit} link and a LR of $10^{-4}$. The optimal configuration depends on the metric we are analysing. In this case, as we are working with an ordinal problem, the most reliable metric is the QWK. However, the rest of the metrics are also included to allow further comparisons with future works.

\subsection{Adience}
\label{sect:adience}
Test results for the experiments made with the Adience dataset are shown in Table \ref{table:AdienceTest}. The best result for each metric is marked in bold and the second best is in italic font.

\begin{table*}[!t]
	\caption{Adience test results. BS stands for Batch Size, LF for link function and LR for Learning Rate. Mean and standard deviation $\text{Mean}_\text{SD}$.}
	\label{table:AdienceTest}
	\footnotesize
	\centering
	\resizebox{\linewidth}{!}{
	\begin{tabular}{c@{\hskip 0.15cm}c@{\hskip 0.15cm}c@{\hskip 0.15cm}c@{\hskip 0.15cm}c@{\hskip 0.15cm}c@{\hskip 0.15cm}c@{\hskip 0.15cm}c@{\hskip 0.15cm}c@{\hskip 0.15cm}c}
		\hline
		\hline
		BS & LF & LR & $\overline{\text{QWK}}_{{SD}}$ & $\overline{\text{MS}}_{{SD}}$ & $\overline{\text{MAE}}_{{SD}}$ & $\overline{\text{CCR}}_{{SD}}$ & $\overline{\text{Top-2}}_{{SD}}$ & $\overline{\text{Top-3}}_{{SD}}$ & $\overline{\text{1-off}}_{{SD}}$\\\hline
		64 & \texttt{clog-log} & $10^{-2}$ & $0.808_{0.025}$ & $0.086_{0.041}$ & $0.147_{0.008}$ & $0.415_{0.031}$ & $0.677_{0.024}$ & $0.798_{0.036}$ & $0.804_{0.015}$\\
		64 & \texttt{clog-log} & $10^{-3}$ & $0.873_{0.006}$ & $0.144_{0.057}$ & $\mathbf{0.124_{0.003}}$ & $\mathbf{0.519_{0.014}}$ & $\mathit{0.764_{0.010}}$ & $0.861_{0.019}$ & $0.886_{0.006}$\\
		64 & \texttt{clog-log} & $10^{-4}$ & $0.799_{0.010}$ & $0.000_{0.000}$ & $0.174_{0.001}$ & $0.324_{0.015}$ & $0.616_{0.020}$ & $0.795_{0.012}$ & $0.771_{0.014}$\\
		64 & \texttt{logit} & $10^{-2}$ & $0.778_{0.019}$ & $0.074_{0.041}$ & $0.159_{0.006}$ & $0.366_{0.025}$ & $0.636_{0.015}$ & $0.785_{0.010}$ & $0.775_{0.015}$\\
		64 & \texttt{logit} & $10^{-3}$ & $\mathbf{0.881_{0.005}}$ & $\mathit{0.178_{0.023}}$ & $\mathit{0.126_{0.001}}$ & $\mathit{0.518_{0.008}}$ & $\mathbf{0.765_{0.015}}$ & $\mathbf{0.902_{0.005}}$ & $\mathbf{0.894_{0.005}}$\\
		64 & \texttt{logit} & $10^{-4}$ & $0.784_{0.011}$ & $0.000_{0.000}$ & $0.180_{0.001}$ & $0.318_{0.026}$ & $0.621_{0.034}$ & $0.772_{0.024}$ & $0.731_{0.030}$\\
		64 & \texttt{probit} & $10^{-2}$ & $0.836_{0.005}$ & $0.135_{0.021}$ & $0.134_{0.002}$ & $0.468_{0.011}$ & $0.720_{0.009}$ & $0.861_{0.009}$ & $0.829_{0.005}$\\
		64 & \texttt{probit} & $10^{-3}$ & $\mathit{0.874_{0.004}}$ & $0.134_{0.012}$ & $0.126_{0.003}$ & $0.511_{0.014}$ & $0.756_{0.009}$ & $\mathit{0.895_{0.003}}$ & $\mathit{0.889_{0.003}}$\\
		64 & \texttt{probit} & $10^{-4}$ & $0.805_{0.004}$ & $0.000_{0.000}$ & $0.170_{0.001}$ & $0.360_{0.011}$ & $0.653_{0.011}$ & $0.809_{0.009}$ & $0.790_{0.009}$\\
		128 & \texttt{clog-log} & $10^{-2}$ & $0.832_{0.013}$ & $0.123_{0.031}$ & $0.135_{0.004}$ & $0.463_{0.013}$ & $0.705_{0.019}$ & $0.813_{0.025}$ & $0.832_{0.006}$\\
		128 & \texttt{clog-log} & $10^{-3}$ & $0.873_{0.006}$ & $\mathbf{0.185_{0.029}}$ & $0.128_{0.002}$ & $0.513_{0.007}$ & $0.758_{0.008}$ & $0.870_{0.011}$ & $0.880_{0.009}$\\
		128 & \texttt{clog-log} & $10^{-4}$ & $0.659_{0.025}$ & $0.000_{0.000}$ & $0.190_{0.002}$ & $0.235_{0.026}$ & $0.466_{0.031}$ & $0.640_{0.030}$ & $0.536_{0.041}$\\
		128 & \texttt{logit} & $10^{-2}$ & $0.781_{0.041}$ & $0.096_{0.059}$ & $0.153_{0.007}$ & $0.398_{0.031}$ & $0.638_{0.033}$ & $0.790_{0.025}$ & $0.779_{0.020}$\\
		128 & \texttt{logit} & $10^{-3}$ & $0.865_{0.005}$ & $0.127_{0.026}$ & $0.134_{0.001}$ & $0.497_{0.009}$ & $0.754_{0.008}$ & $0.882_{0.009}$ & $0.874_{0.008}$\\
		128 & \texttt{logit} & $10^{-4}$ & $0.586_{0.008}$ & $0.000_{0.000}$ & $0.196_{0.001}$ & $0.192_{0.001}$ & $0.364_{0.060}$ & $0.581_{0.034}$ & $0.396_{0.002}$\\
		128 & \texttt{probit} & $10^{-2}$ & $0.849_{0.005}$ & $0.132_{0.010}$ & $0.131_{0.001}$ & $0.479_{0.004}$ & $0.728_{0.007}$ & $0.854_{0.009}$ & $0.847_{0.007}$\\
		128 & \texttt{probit} & $10^{-3}$ & $0.866_{0.002}$ & $0.124_{0.043}$ & $0.130_{0.002}$ & $0.505_{0.006}$ & $0.750_{0.010}$ & $0.882_{0.004}$ & $0.873_{0.006}$\\
		128 & \texttt{probit} & $10^{-4}$ & $0.718_{0.015}$ & $0.000_{0.000}$ & $0.185_{0.001}$ & $0.300_{0.031}$ & $0.575_{0.015}$ & $0.733_{0.010}$ & $0.640_{0.033}$\\
		256 & \texttt{clog-log} & $10^{-2}$ & $0.853_{0.004}$ & $0.157_{0.024}$ & $0.130_{0.002}$ & $0.485_{0.009}$ & $0.744_{0.006}$ & $0.842_{0.016}$ & $0.858_{0.004}$\\
		256 & \texttt{clog-log} & $10^{-3}$ & $0.840_{0.017}$ & $0.095_{0.017}$ & $0.144_{0.005}$ & $0.456_{0.021}$ & $0.720_{0.022}$ & $0.840_{0.018}$ & $0.842_{0.018}$\\
		256 & \texttt{clog-log} & $10^{-4}$ & $0.552_{0.010}$ & $0.000_{0.000}$ & $0.199_{0.001}$ & $0.187_{0.001}$ & $0.368_{0.022}$ & $0.475_{0.025}$ & $0.387_{0.001}$\\
		256 & \texttt{logit} & $10^{-2}$ & $0.764_{0.102}$ & $0.077_{0.067}$ & $0.155_{0.020}$ & $0.387_{0.083}$ & $0.632_{0.103}$ & $0.790_{0.077}$ & $0.783_{0.065}$\\
		256 & \texttt{logit} & $10^{-3}$ & $0.851_{0.008}$ & $0.100_{0.030}$ & $0.147_{0.003}$ & $0.449_{0.015}$ & $0.726_{0.015}$ & $0.861_{0.006}$ & $0.850_{0.008}$\\
		256 & \texttt{logit} & $10^{-4}$ & $0.558_{0.008}$ & $0.000_{0.000}$ & $0.202_{0.001}$ & $0.187_{0.002}$ & $0.206_{0.007}$ & $0.395_{0.046}$ & $0.389_{0.003}$\\
		256 & \texttt{probit} & $10^{-2}$ & $0.858_{0.005}$ & $0.164_{0.033}$ & $0.130_{0.002}$ & $0.486_{0.007}$ & $0.741_{0.008}$ & $0.867_{0.008}$ & $0.862_{0.005}$\\
		256 & \texttt{probit} & $10^{-3}$ & $0.850_{0.008}$ & $0.111_{0.040}$ & $0.144_{0.002}$ & $0.460_{0.011}$ & $0.732_{0.006}$ & $0.865_{0.006}$ & $0.853_{0.007}$\\
		256 & \texttt{probit} & $10^{-4}$ & $0.565_{0.010}$ & $0.000_{0.000}$ & $0.196_{0.001}$ & $0.189_{0.001}$ & $0.409_{0.014}$ & $0.602_{0.022}$ & $0.392_{0.002}$\\
		\hline
		\hline
	\end{tabular}}
\end{table*}

The best mean QWK value was obtained with the \texttt{logit} link function using a BS of 64 and a LR of $10^{-4}$. Also, this configuration obtained the best score for Top-2, Top-3 and 1-off accuracy, and the second best for MS, MAE and CCR. In this case, this configuration can be selected as the optimal for this problem.

\subsection{Statistical analysis}
\label{sect:statisticalanalysis}
In this subsection, a statistical analysis will be performed in order to obtain conclusions from the results. The significance and relative importance of the parameters concerning the results obtained, as well as the most suitable values, were obtained using an ANalysis Of the VAriance (ANOVA).

The ANOVA test~\cite{miller1997beyond} is one of the most widely used statistical techniques. ANOVA is essentially a method of analysing the variance to which a response is subject into its various components, corresponding to the sources of variation which can be identified. ANOVA, in this case, examines the effects of three quantitative variables (termed factors) on one quantitative response. Considered factors are the LF, the LR for the Adam optimization algorithm, and the BS. We assume that five executions are enough to do the statistical tests because of the computational time limitations.

The ANOVA test results show that there are significant differences in average QWK depending on the LF and also depending on the LR for $\alpha=0.05$ ($\text{p-value} = 0.000$). Moreover, an interaction between the LF and the LR can be recognised ($\text{p-value} = 0.001$).


Given that there exist significant differences between the means, we analyse now these differences. A post-hoc multiple comparison test has been performed on the mean QWK obtained. An HSD Tukey's test \cite{tukey1949comparing} has been selected under the null hypothesis that the variance of the error of the dependent variable is the same between the groups. The results of this test over the test set are shown in Table \ref{table:Tukey}. They show that the best LF is the \texttt{clog-log} but the \texttt{probit} link performance is close to it. Also, the best value for the LR parameter is $10^{-3}$. The BS is not relevant for this dataset with the values considered.


The results of the ANOVA III test for the Adience dataset, first, demonstrate that there exist significant differences in average QWK concerning the three factors ($\text{p-value} = 0.000$). Secondly, we found interactions between all the pairs of factors and between all the three factors together (p-values $0.000$, $0.000$, $0.000$ and $0.001$, respectively).

As we did for the DR dataset, a post-hoc multiple comparison test has been performed on the average QWK obtained for Adience. Under the null hypothesis that the variance of the error of the dependent variable is the same between the groups, the HSD Tukey's test has been applied. The results of this test over the test set are shown in Table \ref{table:Tukey}.

\begin{table}[!t]
	\caption{Tukey's test results for both datasets.}
	\label{table:Tukey}
	\centering
	\footnotesize
	\resizebox{\columnwidth}{!}{
	\begin{tabular}{cccccc}
		\hline\hline
		&                   & \multicolumn{2}{c}{\textbf{DR}} & \multicolumn{2}{c}{\textbf{Adience}} \\ \hline
		LF         &        LF         & Mean diff. &        P-val        & Mean diff. &          P-val           \\ \hline
		\texttt{logit}   &  \texttt{probit}  &  $0.046$   &      $0.000$       &  $-0.002$  &         $0.011$         \\
		& \texttt{clog-log} &  $0.084$   &      $0.000$       &  $0.012$   &         $0.000$         \\
		\texttt{probit}  &  \texttt{logit}   &  $-0.046$  &      $0.000$       &  $0.002$   &         $0.011$         \\
		& \texttt{clog-log} &  $0.038$   &      $0.000$       &  $0.014$   &         $0.248$         \\
		\texttt{clog-log} &  \texttt{logit}   &  $-0.084$  &      $0.000$       &  $-0.012$  &         $0.000$         \\
		&  \texttt{probit}  &  $-0.038$  &      $0.000$       &  $-0.014$  &         $0.248$         \\ \hline\hline
		LR         &        LR         & Mean diff. &        P-val        & Mean diff. &          P-val           \\ \hline
		$10^{-2}$     &     $10^{-3}$     &  $-0.046$  &      $0.000$       &  $-0.073$  &         $0.000$         \\
		&     $10^{-4}$     &  $0.148$   &      $0.000$       &  $-0.044$  &         $0.000$         \\
		$10^{-3}$     &     $10^{-2}$     &  $0.046$   &      $0.000$       &  $0.073$   &         $0.000$         \\
		&     $10^{-4}$     &  $0.194$   &      $0.000$       &  $0.029$   &         $0.023$         \\
		$10^{-4}$     &     $10^{-2}$     &  $-0.148$  &      $0.000$       &  $0.044$   &         $0.000$         \\
		&     $10^{-3}$     &  $-0.194$  &      $0.000$       &  $-0.029$  &         $0.023$         \\ \hline\hline
		BS         &        BS         & Mean diff. &        P-val        & Mean diff. &          P-val           \\ \hline
		64         &        128        &  $-0.041$  &      $0.026$       &     -      &            -            \\
		&        256        &  $-0.027$  &      $0.000$       &     -      &            -            \\
		128        &        64         &  $0.041$   &      $0.026$       &     -      &            -            \\
		&        256        &  $0.014$   &      $0.000$       &     -      &            -            \\
		256        &        64         &  $0.027$   &      $0.000$       &     -      &            -            \\
		&        128        &  $-0.014$  &      $0.000$       &     -      &            -\\\hline\hline
	\end{tabular}}
\end{table}

The results over the test set show that the best LF is the \texttt{logit}, the best LR is $10^{-3}$ and the best BS is 128. However, the interactions between these factors made the configuration that uses a \texttt{logit} link, $\eta=10^{-3}$ and BS of 64, the best configuration. It obtained a mean QWK value of $0.940$ for validation and $0.881$ for test. The same parameters, but using the \texttt{probit} link, achieves the second best result ($0.874$). The standard deviation is very low for both cases.

To sum up, the results showed that the best parameter configuration depends on the problem that is being solved. The \texttt{clog-log} function offers the best results in DR dataset while the \texttt{logit} link is the best option for the Adience dataset. However, the best LR for both datasets were $10^{-3}$. It is recommended to use this value for future datasets. The best BS for DR was 10, while the best value for Adience was 128 (intermediate values considered). Finally, there are more interactions between the three factors for the Adience dataset than for DR. These results highlight the importance of adjusting the hyper-parameters for each problem instead of trying to find an optimal configuration for all the datasets. 

\subsection{Comparison with nominal method and previous works}
\label{sect:NominalComparison}
Once the factor parameters have been studied and selected, experiments are run with the standard cross-entropy loss and the softmax function too in order to prove the performance improvement of considering the ordinality of the problem (QWK loss and the CLM). The evaluation metrics remains the same in order to be able to compare. When considering the nominal version, for the DR dataset, the best mean value of QWK was $0.497$ and was obtained when using a BS of 10 and a LR of $10^{-4}$. In the case of Adience dataset, the highest QWK was $0.787$ and was achieved with a BS of 64 and a LR of $10^{-3}$. There are some parameter configurations where the training process gets stagnated and a very low QWK is obtained. As we saw in Sections \ref{sect:dr} and \ref{sect:adience}, this problem is not found when using the ordinal method. 

These results are included in Table \ref{table:Comparison}, together with the comparison against previous works of the state-of-the-art. All the results are given for the test set, except those from \cite{de2018weighted} (DR dataset), because the authors only provided validation results for $128\times 128$ images (however, validation results are usually better than test results). The results for \cite{beckham2017unimodal} were obtained by reproducing their experiments because they did not provide the results for test set. As can be checked, the proposed ordinal model with the best configuration found in the previous step outperforms all the other alternatives in terms of QWK. 

The performance gain of CLM over the nominal version reaches $16.8\%$ for DR and $11.9\%$ for Adience dataset. The DR dataset obtains a higher performance gain from the ordinal method than the Adience dataset. It seems that the method proposed in this work offers a more significant improvement as the given problem complexity increases. Moreover, when compared against alternative ordinal methods (many of them with a deep structure), CLMs are very competitive, possibly because of the flexibility provided by the threshold model structure (where the threshold of each class is independently adjusted). Our proposed model shows a high score and a very low standard deviation, which proves very high robustness compared to other alternatives.

\begin{table}[!t]
	\caption{Comparison between the best results of the CLM network, a nominal network (using softmax and cross-entropy) and previous works for both datasets.}
	\label{table:Comparison}
	\scriptsize
	\centering
	\def\arraystretch{1.3}
	\resizebox{\columnwidth}{!}{
	\begin{tabular}{lccc}
		\hline\hline
		\multicolumn{4}{c}{\textbf{Diabetic Retinopathy}}\\
		\hline
		Method                                    & $\overline{QWK}_{SD}$ & $\overline{CCR}_{SD}$ & $\overline{\text{1-off}}_{SD}$ \\ \hline
		CLM network (proposal)                          &    $0.582_{0.016}$    &    $0.723_{0.004}$    &        $0.868_{0.002}$         \\
		Nominal network                           &    $0.498_{0.013}$    &    $0.692_{0.014}$    &        $0.854_{0.006}$         \\
		de la Torre et al. \cite{de2018weighted}     &  $0.537_{\text{-}}$   &            -            &                -                 \\
		Nebot et al. \cite{nebot2016diabetic}  &  $0.555_{\text{-}}$   &            -            &                -                 \\
		\hline
		\multicolumn{4}{c}{\textbf{Adience}}\\
		\hline
		Method                                    & $\overline{QWK}_{SD}$ & $\overline{CCR}_{SD}$ & $\overline{\text{1-off}}_{SD}$ \\ \hline
		CLM network (proposal)                          &    $0.881_{0.005}$    &    $0.519_{0.013}$    &        $0.894_{0.005}$         \\
		Nominal network                           &    $0.787_{0.004}$    &    $0.458_{0.008}$    &        $0.800_{0.007}$         \\
		Beckham and Pal \cite{beckham2017unimodal} &    $0.855_{0.012}$    &    $0.467_{0.019}$    &        $0.867_{0.011}$         \\
		Eidinger et al. \cite{eidinger2014age} &            -            &    $0.451_{0.026}$    &        $0.807_{0.011}$         \\
		Chen et al. \cite{chen2016cascaded} &            -            &    $0.529_{0.060}$    &        $0.885_{0.022}$         \\
		Levi and Hassner \cite{levi2015age}         &            -            &    $0.507_{0.051}$    &        $0.847_{0.022}$         \\ 
		M. Duan et al. \cite{duan2018hybrid}         &            -            &    $0.523_{0.057}$    &        -         \\ \hline\hline
	\end{tabular}}
\end{table}

\section{Conclusions}
\label{sect:conclusions} 
This paper introduces a new deep ordinal network based on combining CLM models with a continuous QWK loss function. The proposed model is able to improve the performance of the deep network compared to the equivalent nominal version and other models proposed in previous works. Also, it is able to reduce the chance that the model gets stuck when training with some parameter configurations. We conclude that the optimal values for the different parameters considered are problem-dependant. The results highlight the importance of making an experimental design where all of these parameters are adjusted for each problem. In summary, the most significant contributions of the model proposal are the performance increase, the reduction of the number of parameters configurations that should be tried to find the best one and the prevention from over-fitting and stagnation, which grants our method high stability compared to other alternatives.

As future research, it seems that the design of new generalised link functions could be promising, which could be dynamically adapted to any problem based on a learnable parameter. 

\section*{Acknowledgment}
This work has been partially subsidised by the TIN2017-85887-C2-1-P and TIN2017-90567-REDT projects of the Spanish Ministry of Economy and Competitiveness (MINECO), and FEDER funds of the European Union. Víctor Manuel Vargas's research has been subsidized by the FPU Predoctoral Program of the Spanish Ministry of Science, Innovation and Universities (MCIU), grant reference FPU18/00358.

\section*{References}

\bibliography{main.bib}

\end{document}